\documentclass{article}
\usepackage[preprint]{colm2025_conference}
\usepackage{xurl}
\usepackage{microtype}
\usepackage{booktabs}
\usepackage{tabularx}
\usepackage{xcolor}
\usepackage{graphicx}
\usepackage{enumitem}
\usepackage{amsmath}
\usepackage{amssymb}
\usepackage{xspace}
\usepackage{natbib}
\usepackage{float}
\usepackage{multicol}

\newcommand{\system}{\textsc{COLLEAGUE.SKILL}\xspace}
\newcommand{\runhead}[1]{\noindent{\sffamily\bfseries #1}\quad}
\definecolor{HardBlue}{RGB}{0,45,120}
\definecolor{DeepBlue}{RGB}{23,64,123}
\definecolor{DeepTeal}{RGB}{0,106,103}
\definecolor{DeepViolet}{RGB}{93,65,141}
\definecolor{SoftGray}{RGB}{245,246,248}
\definecolor{Copper}{RGB}{177,92,34}
\definecolor{CopperLight}{RGB}{242,226,208}
\definecolor{WarmPanel}{RGB}{253,248,239}
\definecolor{MintPanel}{RGB}{242,250,243}
\definecolor{PaleBlue}{RGB}{239,247,253}
\definecolor{InkGray}{RGB}{62,57,51}
\usepackage[colorlinks,linkcolor=black,citecolor=blue,urlcolor=HardBlue]{hyperref}
\setlist{nosep,leftmargin=*}

\title{\textsc{COLLEAGUE.SKILL}: Automated AI Skill Generation via\\ Expert Knowledge Distillation}

\author{%
  Tianyi Zhou\thanks{Project co-lead}\quad
  Dongrui Liu\footnotemark[1]\quad
  Leitao Yuan\quad
  Jing Shao\quad
  Xia Hu\\[0.4em]
  Shanghai Artificial Intelligence Laboratory\\[0.4em]
  \{zhoutianyi, liudongrui, yuanleitao, shaojing, huxia\}@pjlab.org.cn
}

\begin{document}
\maketitle

\begin{abstract}
LLM agents are increasingly expected not only to complete isolated tasks, but also to carry bounded representations of human expertise, judgment, and interaction style. Building such person-grounded agents remains difficult because actionable knowledge associated with a person or role is usually embedded in heterogeneous traces rather than written as clean instructions. Existing memory and persona systems capture fragments of this evidence, while skill frameworks provide portable packaging formats; however, there is no end-to-end workflow for distilling these traces into inspectable, correctable, and agent-usable skills. We present \system (\url{https://github.com/titanwings/colleague-skill}), an automated trace-to-skill distillation system for generating person-grounded AI skills via expert knowledge distillation. Given materials from a target person or role, \system produces a versioned skill package with two coordinated tracks: a capability track for practices, mental models, and decision heuristics, and a bounded behavior track for communication style, interaction rules, and correction history. The package can be inspected, invoked, updated through natural-language feedback, rolled back, installed across agent hosts, and optionally prepared for controlled distribution. We describe the artifact contract, generation workflow, correction lifecycle, deployment surface, and domain presets implemented in the open-source system. At the time of writing, the public repository has approximately 18.5k GitHub stars; the gallery lists 215 skills from 165 contributors and more than 100k cumulative stars across listed skill cards. The system illustrates how person-grounded skills can be represented as portable, correctable packages rather than opaque prompts or hidden memories. The colleague setting is our primary and most controllable case, but we instantiate the same distillation-and-packaging paradigm in two additional domains: celebrity/public-figure skills, which rely on public evidence and source boundaries, and relationship skills, which require stronger consent, privacy, and local-control assumptions.
\end{abstract}

\section{Introduction}

The role of LLM agents is shifting from executing isolated instructions toward carrying reusable context about how work and interaction should be performed. In practice, users often want an agent to preserve bounded parts of a person's expertise, memory, or interpersonal style: a teammate's review judgment, a specialist's decision heuristics, a public thinker's mental models, or private interpersonal interaction patterns. Rather than treating this demand as unrestricted person simulation, we frame it as \emph{person-grounded trace-to-skill distillation}: turning traces of a person or role into a constrained artifact that makes useful knowledge, interaction style, and limits of use explicit. This framing does not claim identity replacement, and it treats the generated object as an editable technical artifact rather than as the person.

LLM agents increasingly rely on modular extensions. Tools connect agents to external actions, while skills package domain knowledge, procedures, scripts, and reference materials that can be discovered and loaded on demand. This follows a broader shift from single-prompt assistants toward agents that reason and act through external tools, feedback loops, and configurable interaction patterns \citep{yaoReAct2023,schickToolformer2023,shinnReflexion2023,wuAutoGen2023}. The Agent Skills standard defines a skill as a folder centered on a \texttt{SKILL.md} file with metadata and instructions, optionally accompanied by scripts, references, and assets \citep{agentskills2026}. Claude Code similarly treats skills as reusable capabilities that can be invoked directly or loaded when relevant \citep{claudeskills2026}. The format is therefore beginning to act as a portable capability unit for agents.

What remains under-specified is how such skills should be created when the relevant capability or interaction pattern is not already written as an instruction manual. In practice, person-grounded knowledge is often dispersed across heterogeneous traces: a departing teammate's review standards may appear in code comments, incident notes, and chat decisions; a public figure's reasoning style may be expressed across interviews, speeches, essays, and public decisions; and a relationship skill may depend on private interaction histories whose consent and retention boundaries matter. For LLM agents, the challenge is therefore not only to retrieve these materials, but to distill selected evidence into reusable skill packages whose contents, provenance, correction history, and usage limits remain visible.

\system addresses this question through an automated distillation pipeline. The project name reflects its original colleague setting: when a teammate leaves, their local judgment, review standards, incident heuristics, and communication norms often disappear with them. The implemented system generalizes this idea into a broader person-grounded skill workflow. It treats selected traces as evidence for a portable agent skill rather than as a hidden memory store or a claim to reproduce the person.\footnote{Project repository: \url{https://github.com/titanwings/colleague-skill}. Accessed 2026-05-28.} The system accepts chat logs, work documents, email, screenshots, public research material, subtitles, and lightweight user descriptions, then generates a skill package that can be inspected and installed into agent hosts such as Claude Code, OpenClaw, Codex, and Hermes. The colleague, celebrity/public-figure, and relationship variants reuse this package format under different source, evidence, consent, and distribution assumptions.

We study this problem as \emph{person-grounded skill artifact} construction. The target is not an unrestricted conversational model of an individual, but a bounded package of selected capabilities, mental models, communication constraints, examples, and usage boundaries. In a workplace case, this may be an engineer's API review checklist, incident triage heuristics and escalation thresholds. In a celebrity or public-figure case, it may be a source-grounded reasoning style and mental-model library. In a relationship case, it may be a local representation of interaction patterns that should remain editable and deletable. The output is a versioned package whose contents can be examined, corrected, rolled back, deleted, or shared under user control.

We make four contributions:
\begin{itemize}
  \item We formulate person-grounded trace-to-skill distillation as an artifact problem with explicit portability, inspectability, correctability, composability, and governance requirements.
  \item We present the \system pipeline for distilling heterogeneous human traces into a capability track, a bounded behavior track, metadata, host installers, and version state.
  \item We describe the workflow for collection, skill rendering, multi-host installation, natural-language correction, rollback, and optional gallery distribution, with support for domain presets as extensions of the same mechanism.
  \item We document the open-source deployment, public gallery, and extension presets that turn the artifact format into an externally inspectable distribution surface.
\end{itemize}

\section{Problem Formulation}

We use \emph{person-grounded skill} to describe a skill whose instructions are grounded in evidence about a person or role, while remaining bounded by explicit source, usage, and governance constraints. The colleague setting is the primary instance studied here because work expertise gives the clearest utility target and governance boundary. The broader object, however, is not limited to coworkers: the same artifact form can represent public mental models or private interaction patterns under different evidence and consent assumptions.

We define \emph{person-grounded skill generation} as an artifact problem. Given a lightweight profile $p$, a source scope $c$, and a set of source materials $D=\{d_1,\ldots,d_n\}$, the system produces a skill package:
\[
S = (A, M, L),
\]
where $A$ is a set of generated files, $M$ is machine-readable metadata and installation information, and $L$ is lifecycle state such as version, update time, correction count, and rollback history.

The target is not a hidden model of what a real person would say to every possible prompt. The target is a concrete package that distills selected practices and interaction norms into five operational properties:
\begin{enumerate}
  \item \textbf{Portable}: skills-compatible agents can load the package through ordinary skill mechanisms;
  \item \textbf{Inspectable}: users can read extracted rules, examples, limitations, and metadata before use;
  \item \textbf{Composable}: full, work-only, and persona-only entrypoints can be invoked separately;
  \item \textbf{Correctable}: new evidence or user feedback can update the package while preserving prior state;
  \item \textbf{Governable}: metadata, source boundaries, and disclaimers support deletion, sharing decisions, and safety review.
\end{enumerate}

Different domains instantiate $D$ differently. Colleague skills may include design documents, code-review comments, chat decisions, incident notes, and other work traces. Public-figure skills should favor public first-person evidence and long-form interviews. Relationship skills may contain private traces, making consent and local control part of the technical problem rather than a deployment afterthought.

This formulation gives \system a narrower claim than behavioral cloning. The system does not assert that a generated skill is a faithful model of a person. It asserts that selected traces can be transformed into a skills-compatible artifact with explicit files, metadata, entrypoints, correction records, and lifecycle operations. This scope makes the contribution concrete: the artifact can be inspected for structure, source boundaries, update behavior, and deployment compatibility even before downstream human-subject or task-performance studies are available.

\section{\system System Overview}

\begin{figure}[H]
\centering
\makebox[\textwidth][c]{\includegraphics[width=1.05\textwidth]{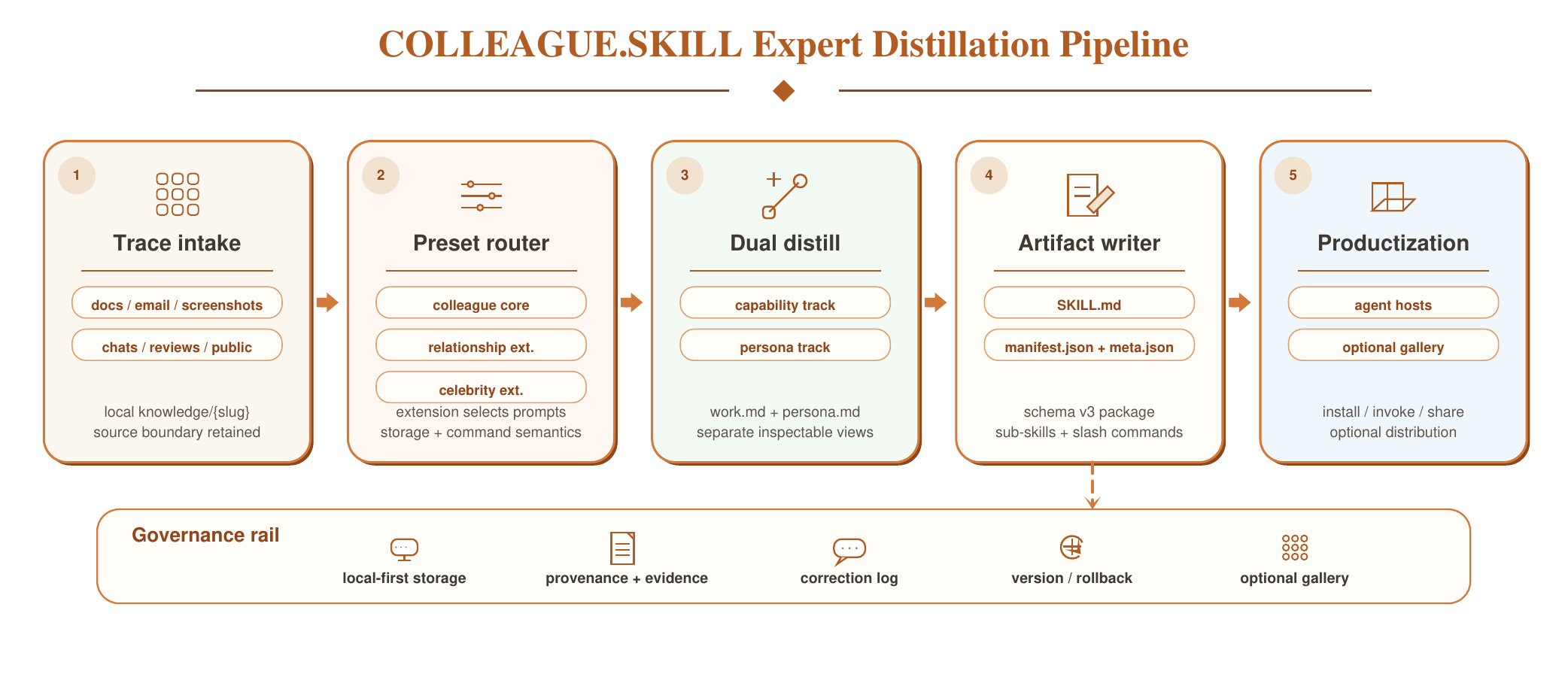}}
\caption{\system architecture for automated person-grounded skill generation. The shared distillation core renders portable agent-skill artifacts; domain presets add source requirements, evidence checks, consent assumptions, and lifecycle or gallery metadata.}
\label{fig:architecture}
\end{figure}

Figure~\ref{fig:architecture} shows the deployed \system architecture. The core path begins with traces of a target person or role: work documents and review comments for a colleague, public interviews and long-form writings for a public figure, or private interaction records for a relationship preset. Collectors and parsers normalize this material into local knowledge directories. Analyzers extract evidence about durable capability, mental models, and bounded interaction style; builders render structured Markdown; and a shared writer produces the generated skill package. The resulting package can be invoked directly, installed into supported hosts, revised through correction records, or, when source rights and metadata permit, prepared for gallery distribution.

\subsection{Application Presets}

\system keeps \texttt{colleague} as the primary preset because it offers a concrete and socially useful starting point: turning a teammate's practices, standards, and communication norms into an inspectable skill. The implementation also makes the source domain explicit so the same artifact workflow can be reused under different evidence and consent assumptions. The repository currently defines three presets: \texttt{colleague}, \texttt{celebrity}, and \texttt{relationship}. Each preset specifies a source boundary, storage root, command aliases, prompt bundle, and optional research or safety tooling.

\begin{figure}[H]
\centering
\makebox[\textwidth][c]{\includegraphics[width=1.05\textwidth]{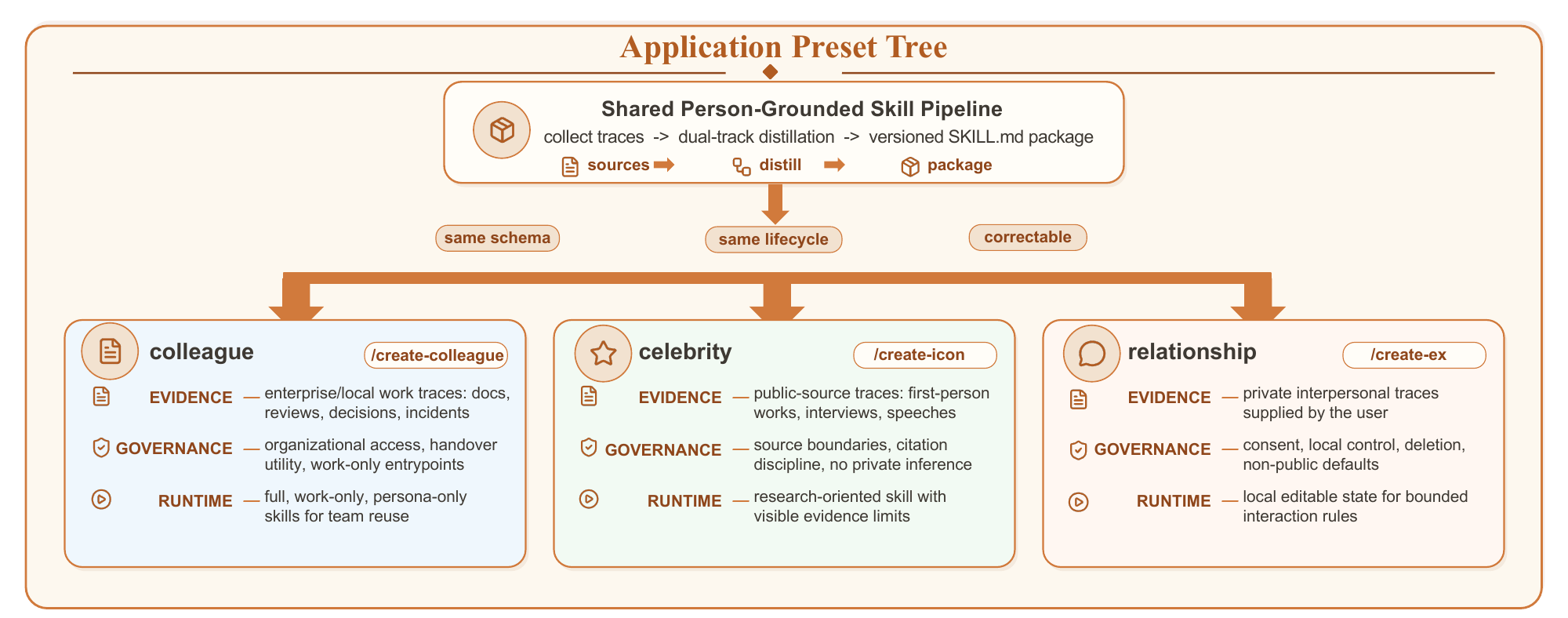}}
\caption{Application presets layered on the \system person-grounded skill pipeline. The shared artifact workflow branches into colleague, celebrity, and relationship presets with different evidence scopes, governance requirements, and invocation aliases.}
\label{fig:presets}
\end{figure}

These presets are domain specializations of the same person-grounded artifact workflow, not separate systems. They avoid duplicating the pipeline when a new application setting needs different prompts, source boundaries, consent defaults, or publication rules. Adding a future preset, such as \texttt{self}, \texttt{author}, or \texttt{team}, then becomes a configuration and prompt-design change rather than a new program.

\subsection{Dual Representation}

Generated \system artifacts use a dual representation. The work or capability track captures responsibilities, workflows, technical standards, review criteria, decision heuristics, and lessons from past work. The implementation names the second track \texttt{persona.md}, but its technical role is narrower: it stores bounded behavior constraints, expression preferences, interaction rules, and correction records. The combined runtime rule is therefore not open-ended impersonation. The agent should select the relevant behavior constraints, apply the capability or mental-model track, and produce a response that remains within the artifact's stated boundaries.

This split is important because many failures in persona systems come from conflating three different things: factual knowledge, procedural judgment, and surface tone. \system makes these pieces inspectable and separately invocable through full, capability-only, and persona-only generated artifacts. In the colleague case, this keeps the main object focused on reusable expert judgment rather than a simulated person; in celebrity/public-figure and relationship presets, the same separation keeps source-grounded mental models or private interaction rules from becoming the system identity.

\subsection{Artifact Schema and Writer}

The writer normalizes metadata into a versioned schema containing identity, preset family, source context, classification, artifact names, engine and toolchain metadata, generation provenance, lifecycle state, and compatibility fields. The current implementation uses schema version 3. It then renders:
\begin{itemize}
  \item \texttt{SKILL.md}: the combined invokable skill;
  \item \texttt{work.md} and \texttt{persona.md}: editable source documents;
  \item \texttt{work\_skill.md} and \texttt{persona\_skill.md}: independently invokable sub-skills;
  \item \texttt{manifest.json} and \texttt{meta.json}: installation, optional gallery, and lifecycle metadata.
\end{itemize}

This is aligned with the Agent Skills standard, where \texttt{SKILL.md} is the required entrypoint and optional files can provide scripts or references \citep{agentskillsspec2026}. The design also follows progressive disclosure: agents see skill metadata first and load detailed instructions only when the skill is invoked \citep{agentskills2026,claudeskills2026}.

The combined \texttt{SKILL.md} contains standard skill frontmatter, including a generated name, a description, and \texttt{user-invocable: true}. Its body embeds the capability track as Part A and the behavior track as Part B. The split entrypoints expose the same tracks independently. This makes the runtime behavior explicit: the artifact can be used as a full person-grounded skill, a capability-only skill, or a behavior-only style reference.

\begin{table}[t]
\centering
\small
\caption{Runtime artifact contract emitted by the shared writer.}
\label{tab:artifact-contract}
\begin{tabularx}{\textwidth}{@{}l l X@{}}
\toprule
Artifact & Primary consumer & Contents \\
\midrule
\texttt{SKILL.md} & Agent runtime, user & Combined invokable skill with frontmatter, capability track, persona track, and operating rules \\
\texttt{work.md} & User, updater & Editable capability document: procedures, standards, heuristics, and task patterns \\
\texttt{persona.md} & User, updater & Editable behavior document: style, interaction posture, boundaries, and correction log \\
\texttt{work\_skill.md} & Agent runtime & Capability-only entrypoint generated from \texttt{work.md} \\
\texttt{persona\_skill.md} & Agent runtime & Persona-only entrypoint generated from \texttt{persona.md} \\
\texttt{manifest.json} & Installers, gallery & Entrypoints, artifact list, compatible runtimes, slash commands, and toolchain metadata \\
\texttt{meta.json} & Lifecycle tools & Schema, provenance, lifecycle version, correction count, and compatibility fields \\
\bottomrule
\end{tabularx}
\end{table}

\section{Generation and Evolution Workflows}

\subsection{Creation Workflow}

Creation begins with the shared person-grounded distillation path: a user provides an alias, optional profile fields, and source material for the target person or role. Repository-supported collectors and import paths cover sources such as Feishu, DingTalk, Slack, WeChat SQLite exports, email archives, PDFs, screenshots, Markdown, and direct paste. Application presets then specialize this creation path. For colleagues, the prompt emphasizes work practice and review judgment. For celebrity/public-figure skills, the \texttt{celebrity} preset adds a research pass over first-person writings, interviews, decisions, expression style, external reception, and timeline evidence. For relationship skills, the \texttt{relationship} preset changes the prompt focus and consent assumptions rather than changing the artifact contract.

The generation prompts then run two conceptual tracks. The capability track extracts durable work methods, expert heuristics, or source-grounded mental models. The behavior track extracts expression and interaction patterns under the preset's boundaries. Builders render structured Markdown, and the writer packages the result into the artifact contract defined above. This separation makes capability and behavior claims inspectable at the file level rather than hiding them in a single prompt.

\subsection{Correction and Update Workflow}

The generated artifact is expected to be imperfect. The correction handler recognizes natural-language feedback such as ``he would not say that'' or ``she would push back here.'' If the correction concerns expert work, it produces a Markdown patch to a relevant section. Patches with matching level-2 headings replace the corresponding section; unmatched sections are appended. If the correction concerns expression or interaction behavior, it produces a normalized correction record:
\[
\{\texttt{scene}, \texttt{wrong}, \texttt{correct}\}.
\]
The writer archives the current version, applies the patch or correction, increments the lifecycle version, and regenerates all derived artifacts. The version manager can list archived versions, back up the current artifacts, roll back to a previous version, and clean old archives.

\begin{figure}[H]
\centering
\makebox[\textwidth][c]{\includegraphics[width=1.05\textwidth]{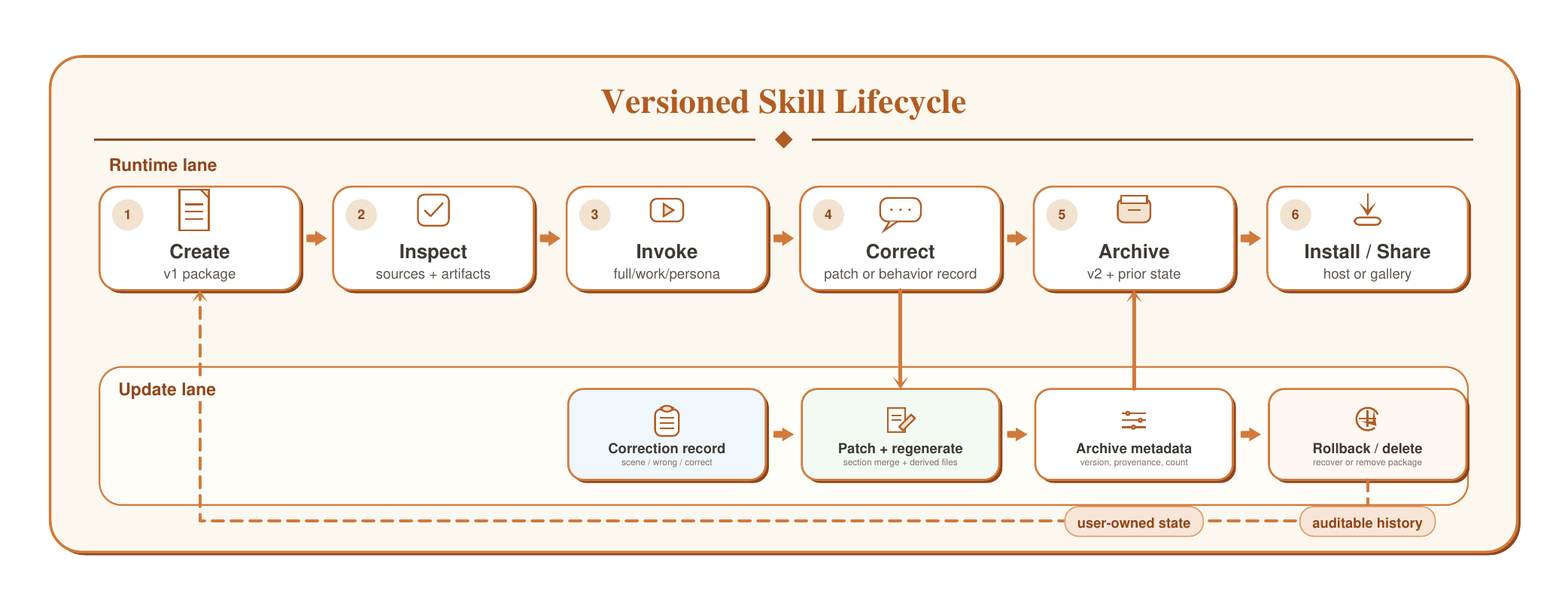}}
\caption{Lifecycle loop for generated skills. Corrections and patches create new versions while preserving rollback points.}
\label{fig:lifecycle}
\end{figure}

\subsection{Public-Figure Research Extension}

The \texttt{celebrity} preset is an extension for public-source expert distillation. Its prompts prioritize first-person works, long-form interviews, documented decisions, and clearly marked inferences over short summaries or content farms. The tooling includes subtitle download, audio transcription, subtitle cleanup, research-note merging, and quality checks. The quality checker scans for mental-model coverage, limitations, expression patterns, internal tensions, grounding URLs, and copyright-safety signals.

The extension makes evidence requirements explicit and executable, but it does not certify factual truth by itself. Instead, it records evidence limits and gives the system a way to downgrade confidence when evidence is thin rather than filling gaps with generic persona text. The research toolchain therefore reuses the same workflow: creation produces inspectable artifacts, correction changes versioned state, and public-facing distribution must expose the evidence limits of the artifact.

\subsection{Relationship Extension}

The \texttt{relationship} preset applies the same artifact workflow to a more sensitive private domain. Its value is not that an agent can replace a person, but that personal interaction traces can be represented as local, editable, and deletable state rather than as an opaque prompt or hidden memory. Compared with the colleague and public-figure settings, this preset requires stronger assumptions about consent, retention, access control, and optional sharing. In the paper's framing, relationship skills stress the governance surface of the package format: they make deletion, correction, local ownership, and non-public defaults first-order artifact requirements.

\section{Deployment and Community Ecosystem}

\system is deployed as an open-source repository with a public site and gallery.\footnote{Project site: \url{https://titanwings.github.io/colleague-skill-site/}. Accessed 2026-05-28.} The site documents the person-grounded skill workflow, installation options, supported sources, and example outputs. The gallery is a downstream sharing layer: generated skills can remain local, be installed into an agent host, or be submitted as shareable packages when the user has rights to publish them. On 2026-05-28, we observed public counters listing 215 skills, 55 meta-skills, and 165 contributors on the gallery, along with repository activity counters. The gallery metadata also records a star count for each skill card; because these counts are synchronized asynchronously and may lag current GitHub state, we report the aggregate at the order-of-magnitude level as more than 100k cumulative gallery stars. We use this statistic only as evidence of public distribution surface, not as adoption quality or task impact.

\begin{figure}[H]
\centering
\makebox[\textwidth][c]{\includegraphics[width=1.05\textwidth,trim=0 34pt 0 12pt,clip]{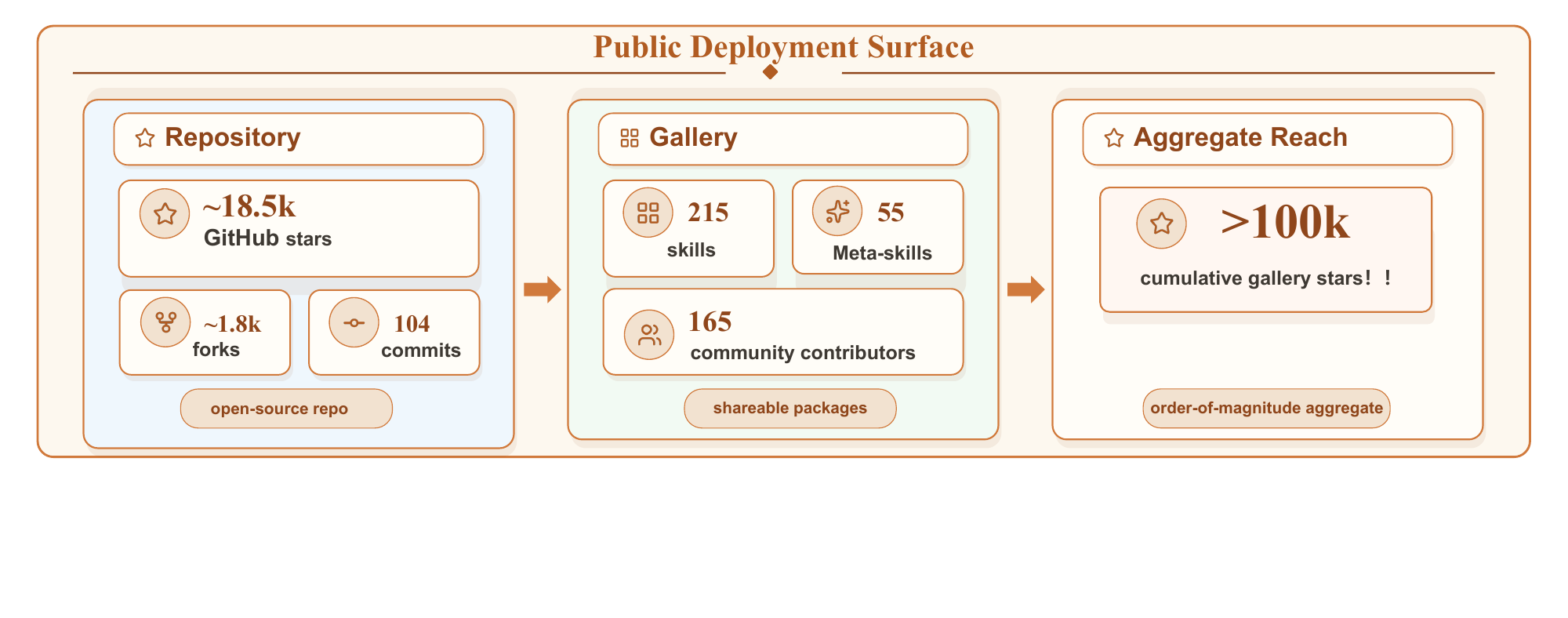}}
\caption{Observed public deployment counters on 2026-05-28. Counts summarize repository activity, gallery scale, and cumulative public signals; they indicate deployment and distribution surface rather than task performance, behavioral fidelity, or adoption-quality metrics.}
\label{fig:deployment}
\end{figure}

The deployment changes the role of \system from a single-prompt construction method to an artifact pipeline for person-grounded skills. Distribution matters because such skills may need to move across hosts, be corrected after use, or be withheld from public sharing. The public site and gallery therefore function as part of the artifact story: they show how generated skills can move from local use to controlled installation and, when appropriate, community sharing beyond the creator's local workspace.

\section{Application Cases}

The following cases show how the shared trace-to-skill workflow appears in different domains. They are design-oriented examples of the artifact workflow, not claims of behavioral equivalence.

\runhead{Colleague skill.} A workplace skill is the most concrete \system instance. It uses private or enterprise material such as design documents, chat decisions, review comments, and incident notes to distill reusable work practice. Its useful behavior is not surface style by itself, but applying review criteria: e.g., checking authentication, input validation, rate limiting, response schema, and sensitive-data exposure before lower-priority issues. The artifact separates these criteria into work rules and behavior constraints, allowing work-only invocation when style transfer is inappropriate.

\runhead{Celebrity skill.} A public-figure skill is an extension built from public evidence. The celebrity family uses a six-dimensional research pass and quality checks to emphasize mental models, citations, and explicit boundaries. A generated skill should indicate where evidence is thin, should not present itself as the actual person, and should remain distinguishable from the workplace case where enterprise traces, access control, and organizational consent dominate the artifact boundary.

\runhead{Relationship skill.} Relationship skills demonstrate the same workflow in a sensitive interpersonal domain. They can represent interaction patterns as local, editable state, but they also expose risks: emotional overattachment, non-consensual simulation, and misuse of private chats. For this family, deployment should prioritize local ownership, deletion, clear disclaimers, and opt-in sharing. We include it as an extension capability, not as an endorsement of unconstrained use.

\section{Related Work}

\runhead{Agent skills and reusable capabilities.} Recent agent systems increasingly externalize capability rather than relying only on monolithic prompts or model weights. ReAct interleaves reasoning traces with actions, Toolformer learns when and how to call APIs, Reflexion and Self-Refine use feedback to revise future behavior, and AutoGen exposes configurable multi-agent conversation patterns \citep{yaoReAct2023,schickToolformer2023,madaanSelfRefine2023,wuAutoGen2023}. AgentBench further shows that agent ability must be evaluated in interactive environments rather than only in static question-answering settings \citep{liuAgentBench2024}. The Agent Skills specification defines a skill as a directory centered on \texttt{SKILL.md}, with optional scripts, references, and assets loaded through progressive disclosure \citep{agentskills2026}. A recent analysis of public Claude skills argues that skills are becoming an infrastructure layer for agents, while also surfacing redundancy, marketplace skew, and safety risks around state-changing actions \citep{lingAgentSkills2026}. \system adopts this emerging package format, but its question is not how to define a skill abstraction. It asks how a person's review standards, mental models, communication constraints, or relationship-specific interaction patterns can be distilled into skills that remain inspectable, editable, portable, and accountable across host environments rather than treated as temporary prompt text or hidden memory.

\runhead{Skill libraries and skill synthesis.} LLM agents have used skill libraries to accumulate reusable behavior. Voyager stores executable code skills in an expanding library and retrieves them to solve new embodied tasks \citep{wangVoyager2023}. More recent systems construct or refine skill knowledge bases from execution trajectories. SkillX distills raw agent trajectories into hierarchical strategic, functional, and atomic skills and refines them through execution feedback \citep{skillx2026}. SkillGen synthesizes auditable skills from successful and failed trajectories and evaluates skills as interventions that can both repair failures and introduce regressions \citep{maSkillGen2026}. AutoSkill abstracts reusable skills from dialogue and interaction traces to support lifelong personalized agents \citep{yangAutoSkill2026}. \system instead distills human traces into person-grounded skills that deliberately separate capability from bounded behavior, expose correction and rollback state, and target installation across multiple agent hosts and sharing surfaces.

\runhead{Memory, personalization, and role-playing agents.} Memory and personalization provide continuity, but they usually keep the representation inside retrieval stores, context managers, or model behavior. Retrieval-augmented generation connects parametric generation with non-parametric memory \citep{lewisRAG2020}. Personalization benchmarks and agent frameworks such as LaMP and PersonaAgent study how models adapt to user histories, preferences, and personalized action spaces \citep{salemiLaMP2024,zhangPersonaAgent2025}. Character-LLM trains role-playing agents from profiles and experiences, RoleLLM constructs role profiles and benchmarks character-level role-playing ability, and SOTOPIA evaluates social intelligence in interactive role-play scenarios \citep{shaoCharacterLLM2023,wangRoleLLM2024,zhouSotopia2024}. \system is deliberately narrower than both traditions: it constructs explicit, reviewable person-grounded skill artifacts that encode selected rules, communication constraints, mental models, limitations, and correction history.

\section{Discussion}

\runhead{Grounded traces, not identity replacement.} \system's core insight is that parts of a person's knowledge, judgment, and interaction style can be distilled into an inspectable AI skill without claiming to reproduce the person. The useful target is a bounded person-grounded artifact: how a person or role weighs evidence, detects risk, explains trade-offs, refuses bad requests, adapts communication to context, or follows documented interaction rules. In the workplace setting, this may be review checklists and incident heuristics; in celebrity/public-figure settings, mental models and cited reasoning patterns; in relationship settings, private interaction constraints under local control. Surface style can make the skill easier to use, but the primary contribution is distilling selected human traces into files that can be inspected, corrected, versioned, installed, and bounded against identity replacement.

\runhead{Why the workflow matters.} A single prompt can mimic surface behavior, but it rarely makes the extracted person-grounded knowledge accountable. \system treats trace-to-skill distillation as a workflow over files: creation, inspection, invocation, correction, rollback, deletion, host installation, and optional distribution. These operations are not auxiliary engineering details. They are the conditions under which a generated person-grounded skill can be audited, repaired, withheld, or shared. They also make the research object sharper. Extraction quality can be inspected at the level of \texttt{work.md} and \texttt{persona.md}; installation and sharing can operate through manifests rather than ad hoc instructions; and governance can operate on explicit metadata rather than hidden prompt state.

The colleague, celebrity, relationship, and gallery work should therefore be read as instances of the same person-grounded skill thesis. Colleague skills test practical expert-knowledge transfer. Celebrity/public-figure skills test whether public mental-model evidence can be packaged with source boundaries rather than becoming generic impersonation. Relationship skills test whether the same artifact controls can protect sensitive private traces. The gallery tests whether generated skills can become a governed distribution layer rather than a private prompt collection. Across these instances, the broader implication is an ecosystem of reusable person-grounded artifacts rather than a gallery of unbounded person simulations: security-review skills, product-decision skills, research-mentor skills, public-thinker mental-model skills, or private interaction skills whose evidence and limits remain visible across installation contexts.

\runhead{Behavioral fidelity frontier.} The claims in this paper are artifact-level claims: \system defines a package format, implements a generation and update workflow, exposes correction and rollback state, supports multiple agent hosts, and demonstrates that the same mechanism can cover colleague, celebrity/public-figure, relationship, and gallery distribution settings. It does not claim that generated skills faithfully reproduce a person or improve downstream work. Those questions require human and task-based studies: whether colleague skills catch the same review issues as the source expert, whether capability-only variants preserve utility without persona risk, whether relationship skills encourage overattachment, whether corrections improve behavior without regressions, and whether public-figure extensions cite evidence rather than hallucinating motives. A useful evaluation protocol should compare full, capability-only, and behavior-only artifacts under matched source evidence, since each variant exposes a different risk-utility trade-off.

\runhead{Productization as a research constraint.} The product surface is part of the research contribution because person-grounded skills become consequential only when they can move across tools, teams, and sharing contexts. Installers, manifests, gallery metadata, rollback state, and deletion paths make the artifact legible to users and hosts rather than leaving it as a private prompt. They also create concrete handles for future study: researchers can compare source scopes, correction records, invocation modes, and publication labels without reverse-engineering hidden memory. In this sense, productization is not a cosmetic layer on top of distillation. It is what turns person-grounded distillation into an inspectable software object whose ownership, provenance, versioning, deployment boundaries, and evaluation handles can be compared, audited, and contested in concrete deployment settings rather than inferred from hidden model behavior.

\section{Limitations and Responsible Deployment}

\system treats person-grounded skills as editable artifacts, not faithful simulations, identity substitutes, or consent proxies. This paper documents the artifact format, workflow, implementation, and public deployment surface, while leaving source matching, task performance, emotional safety, and user trust calibration open. Real deployments will depend on source quality, extraction quality, model behavior, and human review. Corrections can improve an artifact over time, but they can also encode editor bias or make contested traces appear more settled than they are.

Responsible deployment therefore requires explicit participation, scoped source collection, access controls, retention limits, and non-mandatory use. The local-first, inspectable, and versioned design provides useful governance affordances, while lawful source use, consent, and full redaction require separate review. Gallery publication should remain opt-in, with submitter attestation, review, takedown, source-boundary labels, and visible disclaimers for celebrity/public-figure or relationship extensions.

\section{Conclusion}

This paper presented \textsc{COLLEAGUE.SKILL}: Automated AI Skill Generation via Expert Knowledge Distillation. The central claim is not that agents should recreate people, but that selected human traces can be distilled into portable, inspectable skills that encode capabilities, mental models, behavior constraints, and correction history. The colleague setting remains the most concrete starting point, while celebrity, relationship, and gallery components show how the same artifact model extends to broader person-grounded distillation scenarios. The practical message is that digital distillation should produce artifacts that users can read, revise, install, withhold, and delete, rather than opaque prompts that merely sound like a target person. This keeps the paper's ambition grounded: the system does not solve behavioral fidelity, but it makes person-grounded distillation visible enough to be governed, improved, and evaluated. More broadly, \system points toward a product-oriented research path for digital doubles: bounded packages with explicit evidence, rights, correction semantics, and distribution choices. That framing also clarifies what future benchmarks should test: not open-ended impersonation, but whether a bounded package preserves useful judgment while making provenance, consent, and failure modes visible to users. Future work should measure useful judgment and interaction quality without obscuring source quality, consent, provenance, and safety boundaries across concrete deployment settings, artifact variants, and application domains.

\clearpage
\section*{Acknowledgements}

We especially thank the community members who contributed skills, submitted feedback, and supported the public gallery. Their participation helped turn \system from a colleague-knowledge experiment into a broader publicly deployed person-grounded skill ecosystem. We are grateful for their willingness to engage with, extend, and encourage the idea. The GitHub handles below are listed alphabetically, with numeric handles first.

\begingroup
\scriptsize
\setlength{\columnsep}{1.4em}
\begin{multicols}{3}
\raggedright
0xAlexWu\\
123pyLeo\\
1544501967\\
1sh1ro\\
2559063619\\
Aar0nPB\\
AdeleZhu\\
Adrin\\
agenmod\\
AicbLab\\
aka556\\
alchaincyf\\
AnonBug\\
arould001\\
awecsfgvs\\
baibai2013\\
bankeluilian\\
baojiachen0214\\
Bayson-create\\
BeamusWayne\\
binggandata\\
BiscuitCoder\\
bombers26\\
Bughouse1024\\
ByteRax\\
c0dedance\\
Canding3021\\
cantian-ai\\
ccjincc\\
ceetity\\
Charpup\\
ChrisWu11\\
ClarkYoung-xhs\\
CommitHu502Craft\\
cyber-immortal\\
cyberk1895\\
Cyh29hao\\
dadwadw233\\
daiyanpgg-wq\\
DanZai233\\
davecat\\
Dclef\\
derrickgong87\\
dglijin-oss\\
dull-bird\\
DysonSWang\\
EastZsRoad\\
FANzR-arch\\
Fhui\\
Formangarden524\\
gufenglees\\
guilings\\
Hchen1218\\
heywanrong\\
hotcoffeeshake\\
huaqiang-huang\\
HughYau\\
islanddddddd\\
Jack3582-eng\\
Janlaywss\\
jiangziyan-693\\
JikunR\\
JimmyJiang67\\
jinchenma94\\
kangarooking\\
KingOfLitangDz\\
KKKKhazix\\
KKunkuner\\
leezythu\\
leilei926524-tech\\
liangfeiiiii\\
LijiayuDeng\\
linzzzzzz\\
lipG-waver\\
lisi\\
LittleLittleCloud\\
liuyishou-skill\\
lkysyzxz\\
ly-xxx\\
Lyricus233\\
melonlee\\
miaomiao-offical\\
mickey996icu\\
MIMIFY\\
Ming-H\\
Minksgo\\
minruixu\\
miunasu\\
moismin\\
moyvch\\
NatalieCao323\\
Neko-Suwako\\
nicepkg\\
notdog1998\\
nowork-studio\\
nullurl\\
onism11\\
op563296\\
open-source-zjq\\
OpenDemon\\
otter1101\\
Palind-Rome\\
perkfly\\
prog-le\\
Pronting\\
Ratnachem\\
realteamprinz\\
Ricardo-Vv\\
riceshowerX\\
rottenpen\\
SamadhiFire\\
Schlaflied\\
sherjy\\
smallnest\\
snowyowlmia\\
SonicBotMan\\
TammyTan516\\
TerryTian-tech\\
therealXiaomanChu\\
thtang\\
titanwings\\
tmstack\\
To-Carpe-Diem\\
Tomsawyerhu\\
TOPDzZzz\\
Trailblazer-Aha\\
Trust-000\\
UniversePeak\\
VeniVeci\\
vogtsw\\
voidforall\\
Walshyu\\
walter201230\\
wangwu\\
wdl339\\
weixr18\\
whu125\\
wildbyteai\\
will2025btc\\
WilliamX1019\\
wkbin\\
xiaoheizi8\\
xiaoshiyilangzhao1996-droid\\
xr843\\
yangdongchen66-boop\\
yanghaoraneve\\
yaofeino1\\
ybq22\\
yeasy1003\\
yhz-2134\\
YIKUAIBANZI\\
Yinmu\\
YixiaJack\\
Ylsssq926\\
YourongZhou\\
YuzeHao2023\\
z969081067-commits\\
zesion21\\
zgjq\\
zhangeven686-dot\\
zhanghaichao520\\
zhangsan\\
ZhangZangQian\\
Zhrq-vis\\
ZouR-Ma
\end{multicols}
\endgroup

\bibliographystyle{colm2025_conference}
\bibliography{refs}

\end{document}